\documentclass[10pt,twocolumn,letterpaper]{article}

\usepackage{cvpr}
\usepackage{times}
\usepackage{graphicx}
\usepackage{amsmath}
\usepackage{amssymb}
\usepackage{textcomp}
\usepackage{cancel}
\usepackage{xcolor}
\usepackage{array}
\usepackage{tabularx}
\usepackage{stfloats}  
\usepackage{multirow}

\usepackage[pagebackref=true,breaklinks=true,letterpaper=true,colorlinks,bookmarks=false]{hyperref}

\cvprfinalcopy 

\ifcvprfinal\fi
\begin{document}

\title{Mesoscopic photogrammetry with an unstabilized phone camera}

\author{Kevin C. Zhou,* Colin Cooke, Jaehee Park, Ruobing Qian, \\
Roarke Horstmeyer, Joseph A. Izatt, Sina Farsiu \\

Duke University, Durham, NC\\
*{\tt\small kevin.zhou@duke.edu}
}

\maketitle

\begin{abstract}
   We present a feature-free photogrammetric technique that enables quantitative 3D mesoscopic (mm-scale height variation) imaging with tens-of-micron accuracy from sequences of images acquired by a smartphone at close range (several cm) under freehand motion without additional hardware. Our end-to-end, pixel-intensity-based approach jointly registers and stitches all the images by estimating a coaligned height map, which acts as a pixel-wise radial deformation field that orthorectifies each camera image to allow homographic registration. The height maps themselves are reparameterized as the output of an untrained encoder-decoder convolutional neural network (CNN) with the raw camera images as the input, which effectively removes many reconstruction artifacts. Our method also jointly estimates both the camera's dynamic 6D pose and its distortion using a nonparametric model, the latter of which is especially important in mesoscopic applications when using cameras not designed for imaging at short working distances, such as smartphone cameras. We also propose strategies for reducing computation time and memory, applicable to other multi-frame registration problems. Finally, we demonstrate our method using sequences of multi-megapixel images captured by an unstabilized smartphone on a variety of samples (e.g., painting brushstrokes, circuit board, seeds).

\end{abstract}

\begin{figure}
    \centering
    \includegraphics[width=\linewidth]{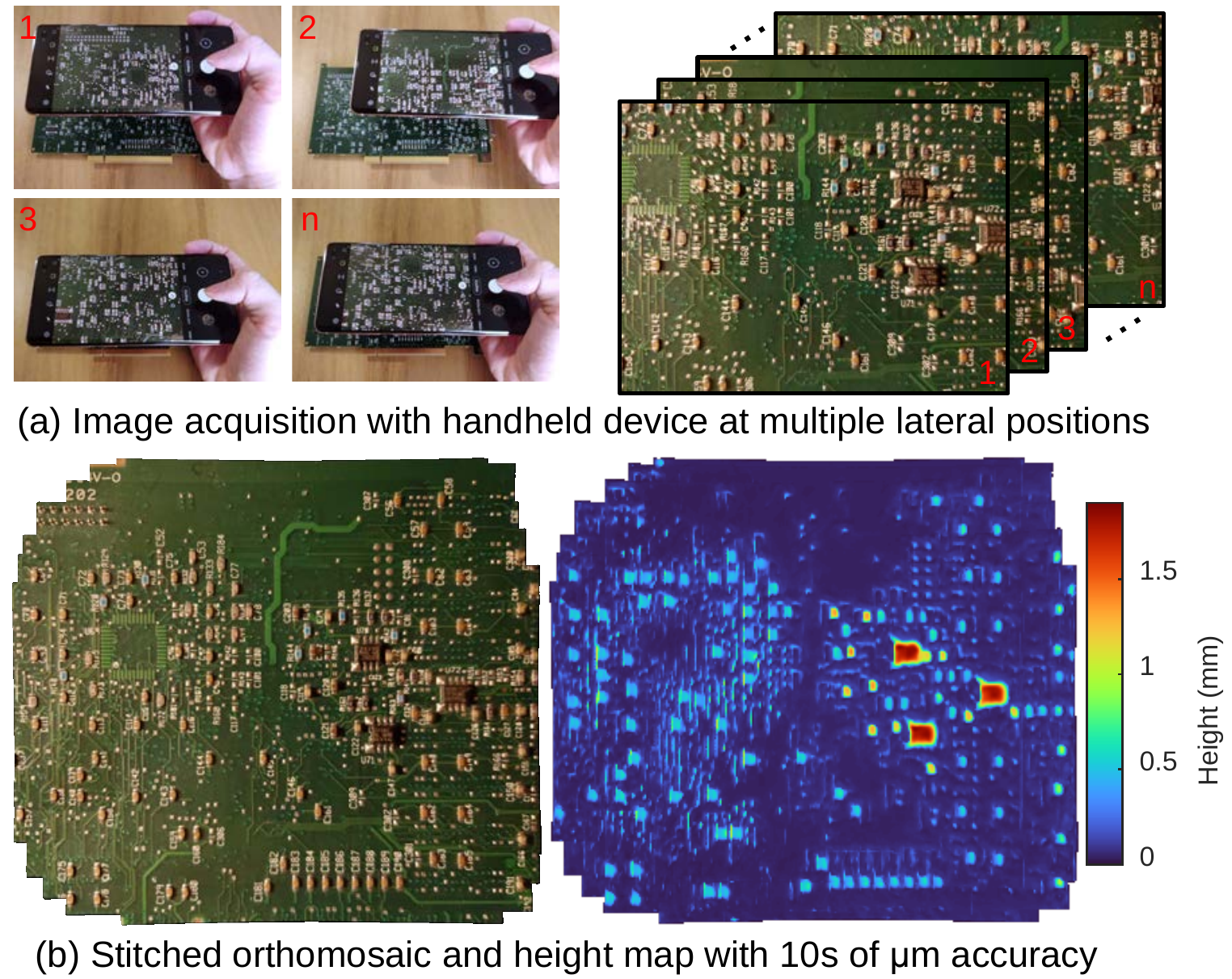}
    \caption{Our method jointly stitches multi-megapixel images acquired under freehand motion at close range and reconstructs high-accuracy height maps without precalibration of camera distortion.}
    \label{fig:teaser}
\end{figure}
\section{Introduction}
The photogrammetric problem of reconstructing 3D representations of an object or scene from 2D images taken from multiple viewpoints is common and well studied, featuring prominently in techniques such as multi-view stereo (MVS) \cite{furukawa2015multi}, structure from motion (SfM) \cite{ullman1979interpretation, wu2013towards,schonberger2016structure}, and simultaneous localization and mapping (SLAM) \cite{fuentes2015visual}. Implicit in these 3D reconstructions is knowledge of the camera parameters, such as camera position, orientation, and distortions, which are jointly estimated in SfM and SLAM. Photogrammetry tools have been developed and applied to both long-range, macro-scale applications \cite{peggs2009recent}, such as building-scale reconstructions or aerial topographical mapping, and close-range, meter-scale applications \cite{luhmann2010close}, such as industrial metrology. However, comparatively less work has been done to push photogrammetry to mesoscopic (mm variation) and microscopic scales, where additional issues arise, such as more limited depths of field and increased impact of camera distortion. Existing approaches at smaller scales typically require very careful camera distortion precalibration, expensive cameras, dedicated setups that allow well-controlled camera or sample motion (e.g., with a dedicated rig), or attachment of control points to the object \cite{luhmann2010close}. 

Here, we show that a smartphone is capable of obtaining quantitative 3D mesoscopic images of objects with 100 \textmu m- to mm-scale height variations at tens-of-micron accuracies with unstabilized, freehand motion and without precalibration of camera distortion (Fig. \ref{fig:teaser}). To achieve this, we present a new photogrammetric reconstruction algorithm that simultaneously stitches the multi-perspective images after warping to a common reference frame, reconstructs sample's 3D height profile, and estimates the camera's position, orientation, and distortion (via a piecewise linear, nonparametric model) in an end-to-end fashion without relying on feature point extraction and matching. We emphasize that our careful modelling of distortions is especially important for mesoscopic applications.  Our method also features a reparameterization of the camera-centric height maps as the outputs of a single \textit{untrained} convolutional neural network (CNN) with the raw camera images at the input (akin to the deep image prior (DIP) \cite{ulyanov2018deep}), which is optimized instead of the height map itself. Since the camera-centric height maps are by design coaligned with the camera images, they are automatically registered once the camera images are registered. As we will demonstrate, both the use of an untrained CNN and careful modeling of distortion substantially reduce reconstruction artifacts, thus allowing high-accuracy height map estimation without camera precalibration or stabilization.

\section{Related work}
The majority of general-purpose implementations of SfM, MVS, and photogrammetry, as well as for more general image registration and stitching problems are centered around feature points \cite{wu2013towards, fuhrmann2014mve, schonberger2016structure, moulon2016openmvg, rupnik2017micmac}. A typical multi-step pipeline starts with extraction of feature points, such as scale-invariant features (SIFT \cite{lowe2004distinctive}), which are then matched across multiple images. The pipeline then crudely estimates the 3D point cloud of the object along with the cameras' positions and poses, which are then refined with bundle adjustment (BA). Our method differs from BA in that it is based on pixel-wise image registration without requiring inference of point correspondences, and operates on rasterized 2D height rather than 3D point clouds, making it more amenable to incorporation of CNNs.

Prior to the development of robust feature point descriptors, pixel-intensity-based techniques for depth estimation from image sequences were also common \cite{matthies1988incremental, horn1988direct, matthies1989kalman, heel1990direct, hanna1991direct, stein2000model, oliensis2000direct}. Our method thus falls in this category. In more recent years, there has been a resurgence of interest in direct approaches with the advent of deep learning. In particular, our work is similar to previous self-supervised deep learning methods that train a CNN on consecutive frames of videos to generate camera-centric depth estimates based on dense pairwise viewpoint warping \cite{zhou2017unsupervised, vijayanarasimhan2017sfm,wang2018learning, li2018undeepvo, godard2019digging, ranjan2019competitive, yin2018geonet,mahjourian2018unsupervised,zhan2018unsupervised, gordon2019depth}. However, they differ in that we jointly warp each frame to a common, potentially unseen reference (e.g., a world reference frame) and stitch them together, while these methods consider consecutive pairs of frames and stay within camera reference frames without stitching to form a larger field of view. Although these methods don't require labels, they still require datasets for training, unlike our method. Other deep learning methods have used multiple frames to estimate depth; however, unlike our method, many of these techniques require known camera poses \cite{wang2018mvdepthnet, huang2018deepmvs, yao2018mvsnet, im2019dpsnet, hou2019multi, murez2020atlas} or reference a keyframe \cite{wang2018mvdepthnet, huang2018deepmvs, yao2018mvsnet, im2019dpsnet, hou2019multi, tang2018ba, teed2018deepv2d, zhou2018deeptam, wei2019deepsfm}, and \textit{all} of these multi-frame methods require supervision and ignore camera distortion. More generally, none of the deep learning approaches mentioned in this paragraph estimate or acknowledge camera distortion, with one exception \cite{gordon2019depth}, which uses a quartic polynomial radial distortion model. However, as we will show, such a model is too limited for obtaining high-accuracy results in high-resolution mesoscopic imaging applications.

While our method uses a CNN, we emphasize that it does not require training on and therefore does not inherit any biases from a dataset. Rather, the CNN serves as a drop-in, compression-based regularizer without the requirement for training or generalization beyond the current sample under investigation. As such, our work is also somewhat similar to recent work on using DIP \cite{ulyanov2018deep} to fill in gaps in camera-centric depth images based on warping to nearby view, which requires knowledge of the camera poses \cite{ghosh2020deep}. While our CNN-based regularizer was inspired by DIP, it differs in that our CNN takes the camera images as input rather than random noise, thus allowing a single shared CNN and therefore a fixed number of parameters regardless of the number of images in the sequence. 

\section{Our approach}
Our new photogrammetric approach is at its core an end-to-end, feature-free, multi-image registration technique formulated as an inverse optimization problem. It is reminiscent of a technique previously developed for jointly registering and combining microscopic images from multiple angles for optical superresolution \cite{zhou2019optical}, and of other pixel-intensity-based multi-image alignment techniques developed for a variety of tasks, such as digital superresolution \cite{robinson2009optimal, aguerrebere2018practical}. Here, our goal is to register and stitch the 2D camera images from multiple views into a single consistent composite mosaic, where the image deformation model is parameterized by the camera model parameters and the sample height map. The key insight of our approach is to use these parameters to warp and co-rectify the camera images so that they appear to have been taken from a single common perspective, thus allowing joint estimation of the stitched RGB image and coaligned height map.

In the following subsections, we describe in detail the camera model we employed to accurately account for distortion and smartphone autofocusing, the perspective rectification models, the multi-image registration framework, and the CNN-based regularization framework. We also propose strategies for dealing with large, multi-megapixel image that alleviate time and memory costs to produce larger, high-resolution RGBH reconstructions (H = height).

\begin{figure*}
    \centering
    \includegraphics[width=\textwidth]{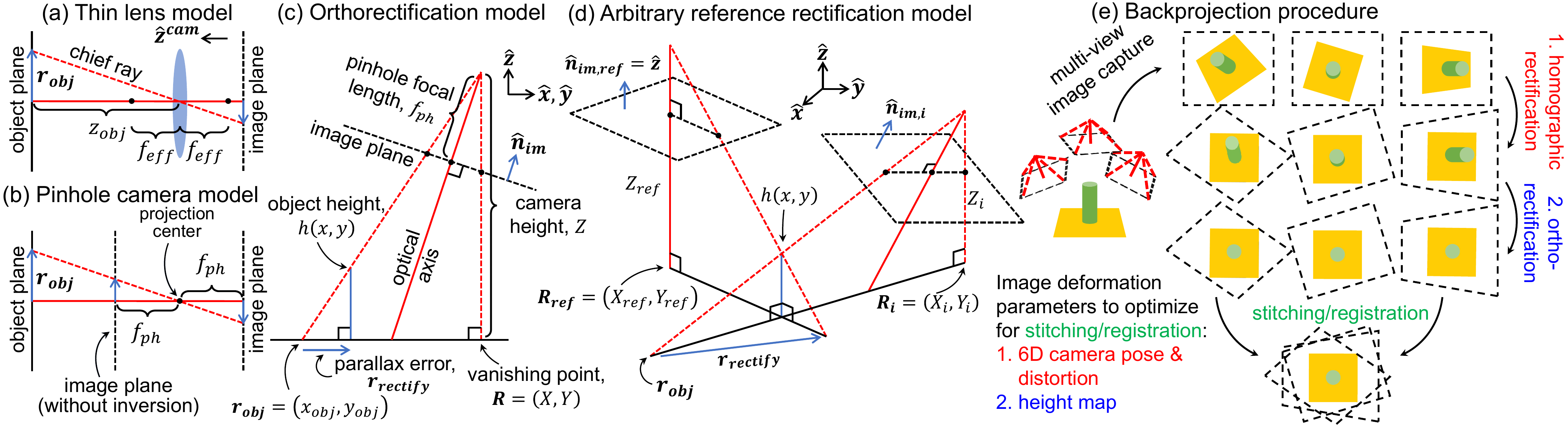}
    \caption{Components of the image deformation model. The chief rays of the thin lens model (a) and the pinhole camera model (b) are equivalent, but have differing definitions of focal length. In general, the scene is not flat (height variation, $h(x,y)$) and the camera is tilted (orientation $\mathbf{\hat{n}_{im}}$). The orthorectification model (c) assumes a reference frame whose projection center is at infinity, such that all projection lines are parallel to $\hat{\mathbf{z}}$. The radial vector field $\mathbf{r_{rectify}}(x,y)$, which converges at the vanishing point ($\mathbf{R}$) and whose magnitudes are $\propto h(x,y)$, performs such a per-pixel rectification, so that the resulting image no longer has perspective distortion. Rectifying the $i^\mathit{th}$ image to an arbitrary reference (b) requires a different $\mathbf{r_{rectify}}$. The backprojection procedure of the camera images (e) to facilitate their stitching/registration requires optimization of the 6D camera pose and distortion, and the height map, which performs orthorectification. See Sec. \ref{model}. }
    \label{fig:camera_model}
\end{figure*}

\subsection{Image deformation model} \label{model}
\noindent
\textbf{Pinhole camera and thin lens model.}
Assuming an ideal camera, image formation can be described by a pinhole camera model, whereby a 3D scene is projected onto a 2D image plane along lines that converge at a point, the center of projection, with a pinhole camera focal length, $f_\mathit{ph}$. The camera itself, however is governed by an effective focal length, $f_\mathit{eff}$, and is related to $f_\mathit{ph}$ by the thin lens equation,
\begin{equation} \label{thin_lens}
    1/z_\mathit{obj}+1/f_\mathit{ph}=1/f_\mathit{eff},
\end{equation}
where $z_\mathit{obj}$ is the distance of the object being imaged from the pinhole camera focus, which corresponds to the position of the thin lens. Thus, the projection lines of the pinhole camera model correspond to chief rays in the lens model (Fig. \ref{fig:camera_model}a,b). When imaging scenes at very far distances ($z_\mathit{obj}\gg f_\mathit{ph}$), $f_\mathit{ph}\approx f_\mathit{eff}$, a good assumption for long-range but less so for very close-range applications.

These models do not predict the limited depths of field associated with closer working distances. To avoid this issue, we assume that the user only attempts to manipulate 2 ($xy$ translation) out of the 6 camera pose parameters, which justifies designating an object plane ($xy$ plane) to which the sample height variations may be referenced, assumed to be within the depth of field. In practice, to account for freehand instability, we still model all 6 degrees of freedom for each image: let the camera's 3D orientation be parameterized by a unit normal vector, $\mathbf{\hat{n}_{im}}$, pointing along the optical axis, and an in-image-plane rotation angle, $\theta$; let the camera's 3D position be designated by the position of its center of projection location, $(X, Y, Z=z_{obj})$. In the supplement, we describe the procedure for homographic rectification of the camera images in a common 2D world reference (Fig. \ref{fig:camera_model}e).

\noindent
\textbf{Camera (un)distortion.}
Camera lenses and sensor placement are not perfect, giving rise to image distortion. This can pose problems for close-range, mesoscale applications, as the 3D-information-encoding parallax shifts become more similar in magnitude to image deformation due to camera distortion. Distortion models commonly separate radial and tangential components, which are often expanded as even-order polynomials \cite{brown1966decentering, fitzgibbon2001simultaneous}. In some cases,the distortion center (the principal point) should also be optimized \cite{hartley2007parameter}. However, as we will show, we found even a 64-order polynomial radial undistortion model insufficiently expressive for our mesoscopic application. Instead, we used a nonparametric, piecewise linear radial undistortion model, whereby the radially dependent relative magnification factor is discretized into $n_r$ points, $\{\widetilde{M}_t\}_{t=0}^{n_r-1}$, spaced by $\delta_r$ with intermediate points linearly interpolated:
\begin{equation}\label{piecewise_linear}
    \widetilde{M}(r)\!=\! 
    \left(1\!+\!\lfloor \frac{r}{\delta_r}\rfloor \!-\!\frac{r}{\delta_r}\right) \widetilde{M}_{\lfloor \frac{r}{\delta_r}\rfloor}
    \!+\!
    \left(\frac{r}{\delta_r}\!-\!\lfloor \frac{r}{\delta_r}\rfloor \right) \widetilde{M}_{\lfloor \frac{r}{\delta_r}\rfloor+1},
\end{equation}
where $\lfloor \cdot \rfloor$ is the flooring operation, $0 \leq r < (n_r-1)\delta_r$ is the radial distance from the distortion center, which is also optimized, and $\widetilde{M}(r)=1$ if there's no distortion. Thus, for a given point in the image, $\mathbf{r_{im}}$, the distortion correction operation is given by
\begin{equation}
    \mathbf{r_{im}} \leftarrow \widetilde{M}(|\mathbf{r_{im}}|)\mathbf{r_{im}},
\end{equation}
which is applied before backprojection. A piecewise linear model, unlike high-order polynomials, also has the advantage of being trivially analytically invertible, allowing easy computation of both image distortion and undistortion. This is important because, while BA typically uses a distortion model, our method requires an \textit{un}distortion model, as we first backproject camera images to form the reconstruction.

\noindent
\textbf{Orthorectification.}
To extend our image deformation model to allow registration of scenes with height variation, we need to warp each backprojected image to a common reference in a pixel-wise fashion. One such option is orthorectification (Fig. \ref{fig:camera_model}c), which can be interpreted as rectifying to a world reference. As such, the effective camera origin is at infinity, so that our images are governed by true length scales, regardless of proximity to the camera (i.e., no perspective distortion). For each camera image backprojection, we estimate a radial deformation field,
\begin{equation} \label{ortho1}
    \mathbf{r_{rectify}}(\mathbf{r_{obj}})=\Delta r\frac{\mathbf{r_{obj}-R}}{|\mathbf{r_{obj}-R}|},
\end{equation}
which is a function of position in the object plane, $\mathbf{r_{obj}}=(x_\mathit{obj},y_\mathit{obj})$, and moves each pixel a signed distance of $\Delta r$ towards the vanishing point, $\mathbf{R}=(X,Y)$, the point to which all lines parallel to $\hat{\mathbf{z}}$ appear to converge in the camera image. $\Delta r$ is itself directly proportional to the height at the new rectified location,
\begin{equation} \label{ortho2}
    h(\mathbf{r_{obj}}+\mathbf{r_{rectify}}) = 
    -Z\frac{\Delta r}{|\mathbf{r_{obj}-R}|}.
\end{equation}
Each camera image has its own height map, forming an augmented RGBH image, which is pixel-wise orthorectified by Eq. \ref{ortho2}.

\noindent
\textbf{Rectification to an arbitrary perspective.}
Orthorectification is just a special case of a more general rectification procedure (Fig. \ref{fig:camera_model}d), where we instead warp to an arbitrary camera-centric reference frame, specified by its vanishing point, $\mathbf{R_{ref}}$, projection center height, $Z_\mathit{ref}$, and an orientation such that the image and object planes are parallel. Given the $i^\mathit{th}$ camera image, whose extrinsics are similarly specified by $\mathbf{R_i}$, $Z_i$, and $\mathbf{\hat{n}_{im,i}}$, the vector that warps a point, $\mathbf{r_{obj}}$, to the reference frame is given by 
\begin{equation}
\begin{gathered}
    \mathbf{r_{rectify}}(\mathbf{r_{obj}})=\\
    \frac{h}{Z_i}\frac{Z_i-Z_\mathit{ref}}{Z_\mathit{ref}-h}(\mathbf{r_{obj}-R_i})
    +
    \frac{h}{Z_\mathit{ref}-h}(\mathbf{R_i}-\mathbf{R_{ref}}).
\end{gathered}
\end{equation}
If we allow $Z_\mathit{ref}\rightarrow\infty$, the result is consistent with Eqs. \ref{ortho1}, \ref{ortho2}, and we recover the orthorectification case, as expected.

\noindent
\textbf{Accounting for autofocus.}
Most smartphones have an autofocus feature where the sensor or lens position automatically adjusts to sharpen some part of the image. This feature is only relevant for close-range applications, as can be seen in Eq.\ref{thin_lens}, and manifests as a dynamically adjusted $f_\mathit{ph,i}$ for the $i^\mathit{th}$ image, while for long-range applications, $f_\mathit{ph,i}$ remains fixed at $f_\mathit{eff}$. This issue is intertwined with the well-known limitation of photogrammetry that inferring absolute scale requires something of known length in the scene (ground control points). In particular, Eq. \ref{ortho2} by itself is insufficient to obtain quantitative height maps, because $Z_i$ and $f_\mathit{ph,i}$ are ambiguous up to a scale factor related to the camera's magnification for the $i^\mathit{th}$ image:
\begin{equation}\label{magnification}
    M_i = f_\mathit{ph,i}/Z_i \approx M_0Z_0/Z_i,
\end{equation}
where the approximation assumes that the camera has autofocused once and maintained approximately the same $f_\mathit{ph}$ throughout data acquisition. The more consistent the height of the camera during acquisition, the better this approximation. The advantage of this approximation is not having to calibrate the magnification for each image in the sequence.
Combining Eqs. \ref{thin_lens}, \ref{ortho2}, and \ref{magnification}, we obtain the ambiguity-free height estimate by the $i^\mathit{th}$ image,
\begin{equation}\label{height_estimate}
\begin{aligned}
    h_i(\mathbf{r_{obj}}+\mathbf{r_{rectify}})
    &\approx -f_\mathit{eff}\frac{\Delta r_i}{|\mathbf{r_{obj}-R_i}|}
    \left(1+\frac{1}{M_0}\frac{Z_i}{Z_0}\right).
\end{aligned}
\end{equation}

\noindent
\textbf{Putting it all together.}
Homographic rectification using the pinhole/thin-lens models, camera undistortion, and orthorectification (or arbitrary-reference rectification) via the height map, as described in this section (\ref{model}), together constitute the backprojection step (Fig. \ref{fig:camera_model}e, \ref{fig:method_diagram}). Let the associated image deformation parameters be collectively denoted as $\mathbf{w}$.

\subsection{Multi-frame image stitching and registration}
Given the current estimate of the image deformation parameters, $\mathbf{w}$, we simultaneously backproject all the images to form an estimate of the RGBH reconstruction, $\mathbf{B}$, with the coaligned height map stacked as the fourth channel: 
\begin{equation}
    \mathbf{B}\leftarrow \mathbf{0},\qquad
    \mathbf{B}[\mathbf{x_w},\mathbf{y_w}]\leftarrow \mathbf{D_{RGBH}},
\end{equation}
where $(\mathbf{x_w,y_w})$ are the flattened coordinates corresponding to the pixels of $\mathbf{D_{RGBH}}$, which are the flattened RGB images augmented with the camera-centric height maps. If a pixel of $\mathbf{B}$ is visited multiple times, the values are averaged.

\begin{figure}
    \centering
    \includegraphics[width=.95\linewidth]{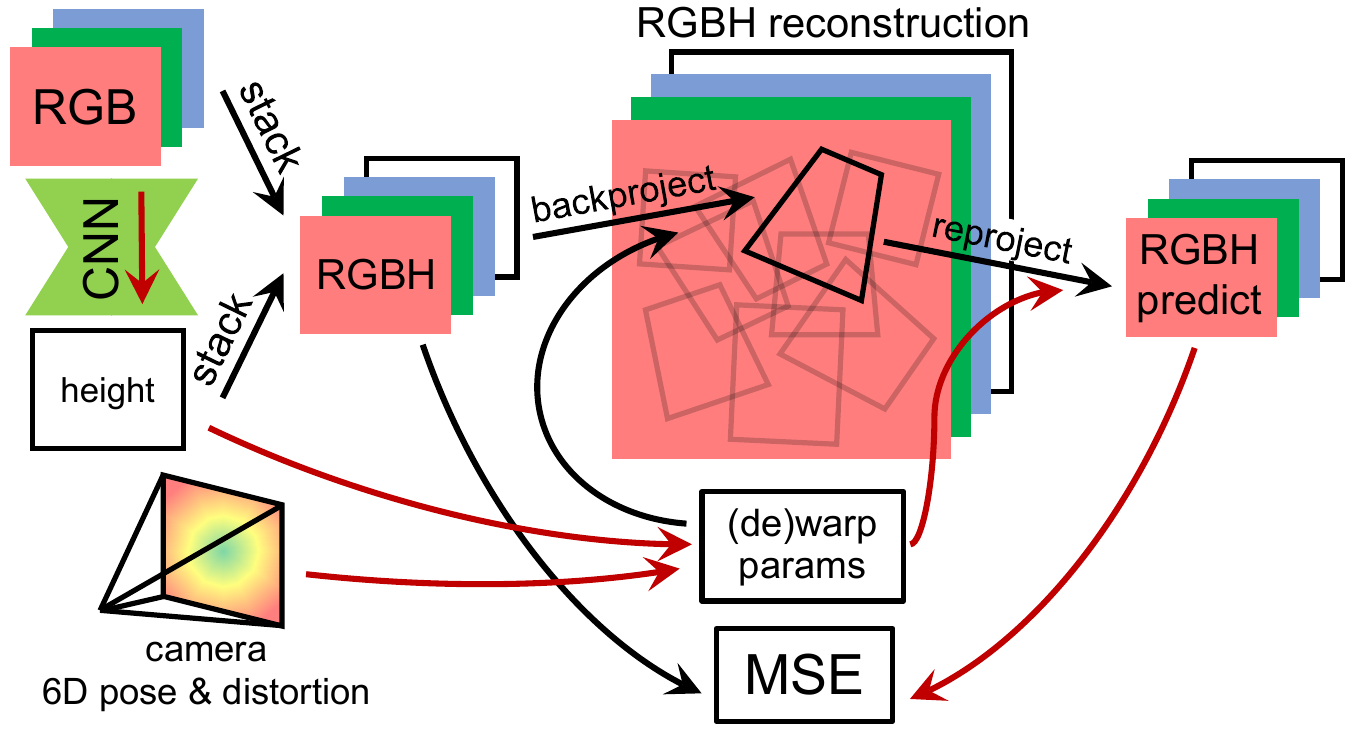}
    \caption{Method overview. A stack of RGB images (only one shown for clarity) is augmented with their camera-centric height maps (RBGH), which are reparameterized as outputs of an untrained CNN. The camera-centric height maps and camera parameters serve as the image deformation parameters, which rectify the RGBH images so that they can be registered and stitched (RGBH reconstruction). We use the same warping coordinates to reproject the images to form a prediction, whose MSE with RGBH is minimized with respect to the CNN and camera parameters. Arrows indicate forward differentiable operations, but only red arrows indicate paths that allow backpropagation (Sec. \ref{stop_gradient}).}
    \label{fig:method_diagram}
\end{figure}

To guide the optimization, we next generate forward predictions of the camera images, $\mathbf{\hat{D}_{RGBH}}$, by using the exact same backprojection coordinates, $(\mathbf{x_w,y_w})$, to \textit{reproject} back into the camera frames of reference and compute the mean square error (MSE) with the original camera images (Fig. \ref{fig:method_diagram}). The idea is that if the backprojected images are consistent with each other at the pixels where they overlap, then the forward predictions will be more accurate. We then use gradient descent to minimize the MSE with respect to the image deformation parameters, that is
\begin{equation}\label{MSE}
    \underset{\mathbf{w}}{\text{min}}\  ||\mathbf{\hat{D}_{RGBH}}-\mathbf{D_{RGBH}}||^2.
\end{equation}

To avoid local minima, we adopted a multi-scale strategy, whereby both $\mathbf{D_{RGBH}}$ and $\textbf{B}$ were subject to a downsampling procedure that was relaxed over time. Further, we didn't update the height map until we reached the lowest downsampling factor. If the scene consisted of non-repetitive structures and the camera images exhibited a lot of overlap, initializing each image to the same position was often a good initial guess. However, if this failed, we initialized using sequential cross-correlation-based estimates.

\subsection{Regularization}
\noindent
\textbf{Reparameterization with a CNN.}
We reparameterized the camera-centric height maps as the output of a CNN with the respective RGB images as the inputs. Instead of optimizing for the per-image height maps, we optimize the weights of a single untrained CNN as a DIP, whose structure alone exhibits a bias towards ``natural'' images, as empirically demonstrated on multiple image-reconstruction-related tasks \cite{ulyanov2018deep}, including 3D reconstruction \cite{zhou2020diffraction}. While the DIP is often an overparameterization, in our case the single CNN has fewer parameters than the total number of height map pixels that we otherwise would have directly optimized, thus offering further means of regularization \cite{heckel2018deep}. Furthermore, we used an encoder-decoder network architecture without skip connections, forcing the information to flow through a bottleneck. Thus, the degree of compression in the CNN is an interpretable regularization hyperparameter, where restricting information flow may force the network to discard artifacts (see supplement for architecture). Finally, we note that the network doesn't need to generalize beyond the current image sequence.

\noindent
\textbf{Camera-centric height map consistency.}
Although we directly optimize for camera-centric height maps, they ultimately follow the same backprojection and reprojection procedure that the RGB images undergo, as described earlier. However, while the contribution of RGB pixels to the MSE in Eq. \ref{MSE} serves as feedback for registration, the contribution of the height values to the MSE is primarily to make the camera-centric height maps more consistent irrespective of the backprojection result. This can be useful, for example, when filling in height values at the vanishing points, which are blind spots, as $h\propto\mathbf{r_{rectify}}(\mathbf{R})=0$. Since RGB values and height values are not directly comparable, we introduce a regularization hyperparameter that scales their relative contributions.

\subsection{Reducing computation time and memory}
To reduce computation time and memory, we used gradient checkpointing \cite{chen2016training, griewank2000algorithm} and CPU memory swapping \cite{le2018tflms}, as well as two novel strategies, which we discuss next.

\noindent
\textbf{Blocking backpropagation through the reconstruction} \label{stop_gradient}
Instead of computing the total gradient of the loss with respect to the image deformation parameters, which would require backpropagation across every path that leads to the deformation parameters, we compute partial gradients using only the paths that lead to the deformation parameters without going through the reconstruction (red arrows in Fig. \ref{fig:method_diagram}).
Writing out the relevant terms in the chain rule expansion of the gradient of the loss with respect to the image deformation parameters (Sec. \ref{model}), we have
\begin{equation}
    \frac{dL}{d\mathbf{w}}=\frac{\partial L}{\partial \mathbf{\hat{D}_{RGBH}}}
    \left(
    \mathbf{J}_{\mathbf{\hat{D}_{RGBH}}}(\mathbf{w}) + 
    \cancelto{0}{\mathbf{J}_{\mathbf{\hat{D}_{RGBH}}}(\mathbf{B})
    \mathbf{J}_\mathbf{B}(\mathbf{w})}
    \right),
\end{equation}
where $\mathbf{J_y(x)}$ denotes the Jacobian of $\mathbf{y}$ with respect to $\mathbf{x}$ and $L$ is the loss. Doing so ends up saving memory and time because it avoids computing expensive derivatives associated with the reconstruction backprojection and reprojection steps. An intuitive interpretation is that at each iteration the reconstruction serves as a temporarily static reference to which all the images are being registered, as opposed to also explicitly registering the reconstruction to the images. Note, however, that both registration directions are governed by the same parameters, and that this ``static'' reference still updates at every iteration.

\begin{figure*}[ht]
    \centering
    \includegraphics[width=.95\textwidth]{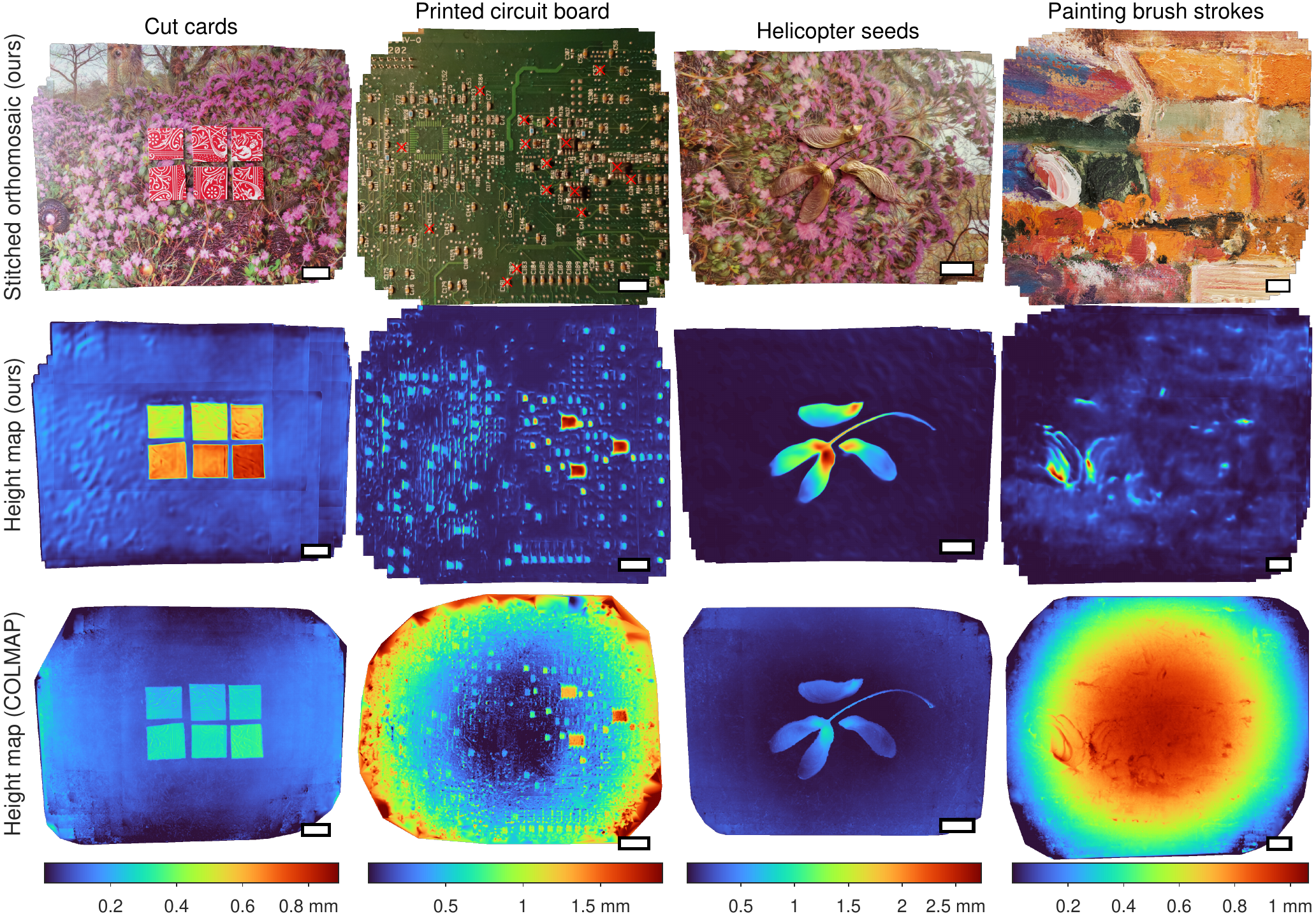}
    \caption{Stitched orthomosaics and height maps for various samples with mm-scale height variation. Height values for the cut cards and PCB components (red \textcolor{red}{$\times$}'s) are quantified in Tables \ref{card_analysis} and \ref{pcb_analysis}, respectively. COLMAP consistently underestimates heights. Scale bars, 1 cm.}
    \label{fig:results}
\end{figure*}

\noindent
\textbf{Batching with a running-average reconstruction.}
At every iteration of the optimization, we require an estimate of the reconstruction, which itself requires joint participation of all images in the dataset to maximize all the available information. This can be problematic because it requires both the reconstruction and the entire dataset to be in GPU memory at the same time. A standard approach to address this problem is batching, which at first glance would only only work for the reprojection step, as the projection step requires all images to form the reconstruction. A two-step approach to overcome this requirement is to realize that, because of our strategy of blocking backpropagation paths through the reconstruction, we can simply generate the reconstruction incrementally in batches without worrying about accumulating gradients. Once the temporarily static reconstruction is generated given the current estimates of the image deformation parameters, we could then update the parameters by registering batches of images to the reconstruction. This approach, however, is inefficient because it requires two passes through the dataset per epoch, and the parameters are only updated during one of the passes.

We introduce a more efficient, one-step, end-to-end strategy where each batch updates \textit{both} the reconstruction \textit{and} the parameters by keeping track of a running average of the reconstruction. In particular, the update rule for the reconstruction after the $(j+1)^\mathit{th}$ gradient step when presented with the $j^\mathit{th}$ batch as a list of warped coordinates and their associated RGB values, $(\mathbf{x}_\mathit{w,j},\mathbf{y}_\mathit{w,j}, \mathbf{D}_j)$, is given by
\begin{equation} \label{batch}
\begin{gathered}
    \mathbf{B}_{j+1}\leftarrow\mathbf{B}_j\\
    \mathbf{B}_{j+1}[\mathbf{x}_\mathit{w,j},\mathbf{y}_\mathit{w,j}]\leftarrow m\mathbf{B}_j[\mathbf{x}_\mathit{w,j},\mathbf{y}_\mathit{w,j}]+(1-m)\mathbf{D}_j
\end{gathered}
\end{equation}
where $0<m<1$ is the momentum controlling how rapidly to update $\textbf{B}$. The batch is specified very generally in Eq. \ref{batch}, and can correspond to any subset of pixels from the dataset, whether grouped by image or chosen from random spatial coordinates. Only the spatial positions of the reconstruction visited by the batch are updated in the backprojection step, and the loss is computed with the same batch after the reprojection step. As a result, we only need one pass though the dataset per epoch. This method is general and can be applied to other multi-image registration problems.

\section{Experiments}
Using the rear wide-angle camera of a Samsung Galaxy S10+ ($f_\mathit{eff}$ = 4.3 mm) and freehand motion, we collected multiple image sequence datasets consisting of 21-23 RGB $1512\times2016$ images ($2\times$-downsampled from $3024\times4032$). While our method does not require it, we attempted to keep the phone approximately parallel and at a constant height (5-10 cm) from the sample while translating the phone laterally, to keep as much of the sample as possible within the limited depth of field associated with such close working distances. To obtain absolute scale, we estimated the magnification of the first image of each sequence using reference points of known separation in the background.

We implemented our algorithm in TensorFlow \cite{abadi2016tensorflow} (code and data available at github.com/kevinczhou/mesoscopic-photogrammetry) and performed reconstructions on an Intel Xeon Silver 4116 processor augmented with an 11-GB GPU (Nvidia RTX 2080 Ti). We used the same CNN architecture for all experiments, tuned on an independent sample to balance resolution and artifact reduction (see supplement for architectures). Gradient descent via Adam \cite{kingma2014adam} was performed for 10,000 iterations for each sample with a batch size of 6. We set $n_r=30$ (Eq. \ref{piecewise_linear}) and $m=0.5$ (Eq. \ref{batch}).

\begin{table}
\begin{tabular}{ c | c c | c c | c c}
 Card \# & \multicolumn{2}{c|}{Ours} & \multicolumn{2}{c|}{CM} & \multicolumn{2}{c}{CM (scaled)}\\
 (G. T.) & Acc. & Prec. & Acc. & Prec. & Acc. & Prec.\\ 
 \hline
 bkgd (0) & 59.4 & 45.5 & 235.6 & 59.6 & 19.2 & 157.8\\  
 \#1 (295) & 34.9 & 37.7 & 74.9 & 26.8 & 41.8 & 71.0\\  
 \#2 (350) & 2.0 & 36.5 & 24.0 & 24.6 & 2.4 & 65.2\\  
 \#3 (420) & 38.5 & 54.9 & 6.7 & 18.4 & 31.7 & 48.9\\  
 \#4 (485) & 6.0 & 26.2 & 55.6 & 20.6 & 9.4 & 54.7\\  
 \#5 (555) & 32.5 & 35.5 & 120.6 & 15.4 & 47.3 & 40.8\\  
 \#6 (625) & 10.5 & 28.4 & 151.6 & 18.1 & 14.1 & 47.9\\  
 \hline
 mean & 26.3 & 37.8 & 95.6 & 26.2 & 23.7 & 69.5
\end{tabular}
\caption{Accuracy (abs. error from ground truth (G. T.)) and precision (st. dev.) of our method vs. COLMAP (CM) vs. CM rescaled to match G. T. of the cut card sample (Fig. \ref{fig:results}). All units are \textmu m.}
\label{card_analysis}
\end{table}

\newcommand{\STAB}[1]{\begin{tabular}{@{}c@{}}#1\end{tabular}}
\begin{table*}[!ht]
\newcolumntype{Y}{>{\centering\arraybackslash}X}
\small
\begin{tabularx}{\linewidth}{@{\hskip3pt}c@{\hskip3pt} p{.5cm} | Y Y Y Y Y Y Y Y Y Y Y Y Y Y Y Y c| Y}
 & & bkgd & C120 & C121 & C126 & C142 & C147 & C181 & C182 & C203 & C57 & R135 & R144 & R175 & R184 & U70 & U71 & U72 & mean\\ 
 \hline
 & G.T. & 0 & 1253 & 1257 & 1269 & 618 & 1282 & 632 & 632 & 1354 & 677 & 533 & 548 & 476 & 427 & 1772 & 1771 & 1778 & --\\  
\multirow{2}{*}{\STAB{\rotatebox[origin=c]{90}{Ours}}}
 & Acc. & 119.8 & 41.4 & 15.8 & 79.0 & 60.1 & 163.6 & 5.8 & 56.0 & 43.5 & 94.3 & 122.4 & 89.3 & 21.0 & 52.7 & 44.3 & 27.5 & 31.9 & 62.9\\  
 & Prec. & -- & 138.4 & 123.5 & 133.2 & 32.2 & 175.1 & 45.3 & 34.2 & 86.7 & 139.2 & 117.0 & 61.0 & 17.6 & 52.9 & 101.1 & 141.6 & 134.3 & 95.8\\  
 \multirow{2}{*}{\STAB{\rotatebox[origin=c]{90}{CM}}}
 & Acc. & 455.7 & 435.4 & 557.7 & 193.9 & 116.6 & 90.7 & 401.6 & 225.0 & 456.9 & 548.0 & 74.6 & 173.3 & 452.5 & 305.4 & 502.1 & 388.9 & 168.3 & 326.3\\
 &  Prec. & -- & 108.0 & 140.2 & 83.4 & 248.1 & 163.4 & 118.5 & 95.5 & 133.8 & 51.7 & 98.7 & 91.3 & 105.0 & 70.4 & 90.9 & 95.3 & 123.6 & 113.6\\ 
\end{tabularx}
\caption{Quantification of accuracy and precision (in \textmu ms) of our method and COLMAP (CM) on PCB components (Fig. \ref{fig:results}, red \textcolor{red}{$\times$}'s).}
\label{pcb_analysis}
\end{table*}

We compare our method to the open-source, feature-based SfM tool, COLMAP \cite{schonberger2016structure}, which has been shown to outperform competing general-purpose SfM tools \cite{bianco2018evaluating}. See supplement for COLMAP hyperparameter settings.

\begin{figure}
    \centering
    \includegraphics[width=\columnwidth]{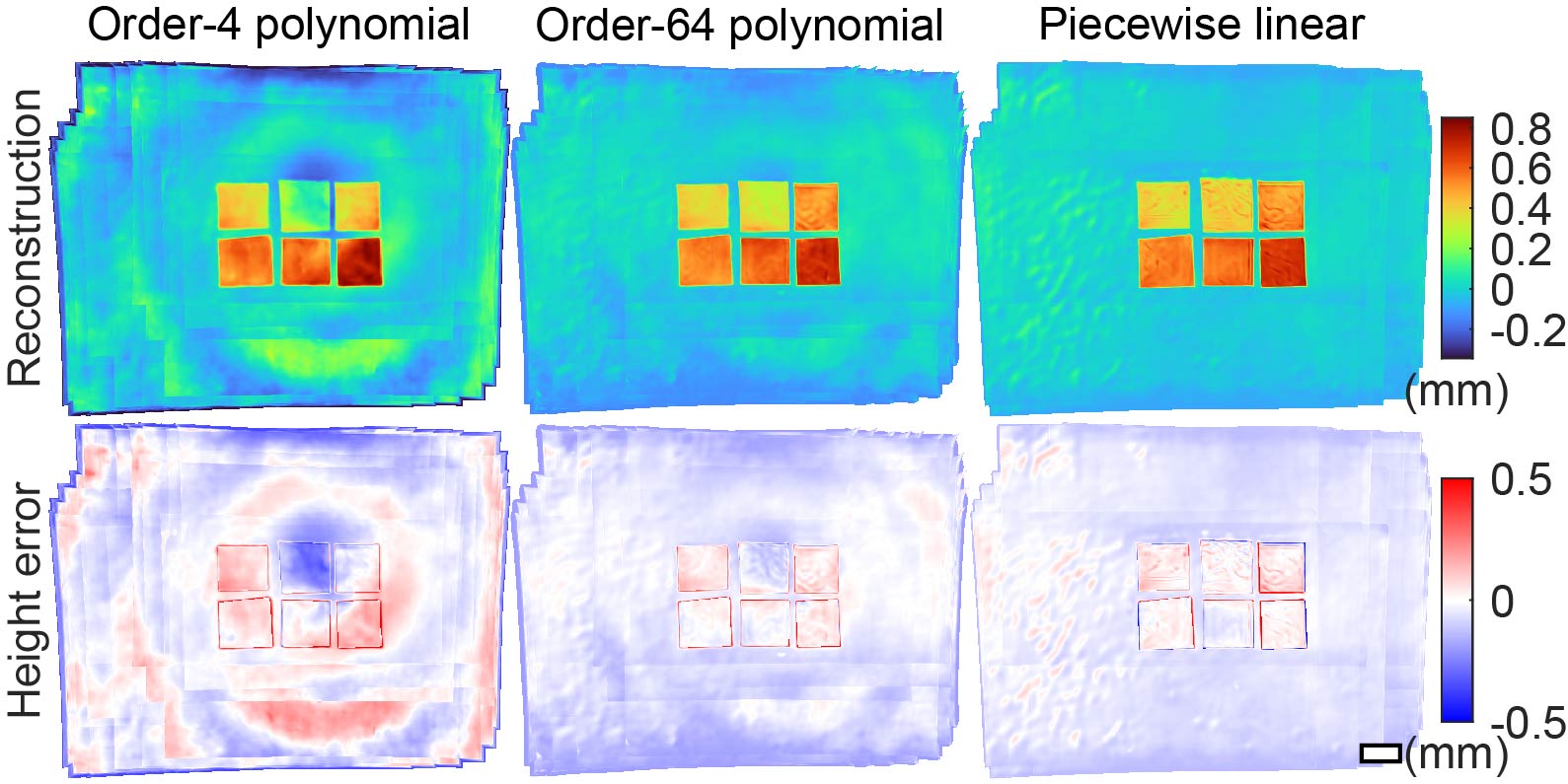}
    \caption{Visual comparison of height map reconstructions of the cut cards sample using different undistortion models. Note the ring artifacts with the polynomial models. Scale bar, 1 cm.}
    \label{fig:distortion_cards}
\end{figure}
\begin{figure}
    \centering
    \includegraphics[width=\columnwidth]{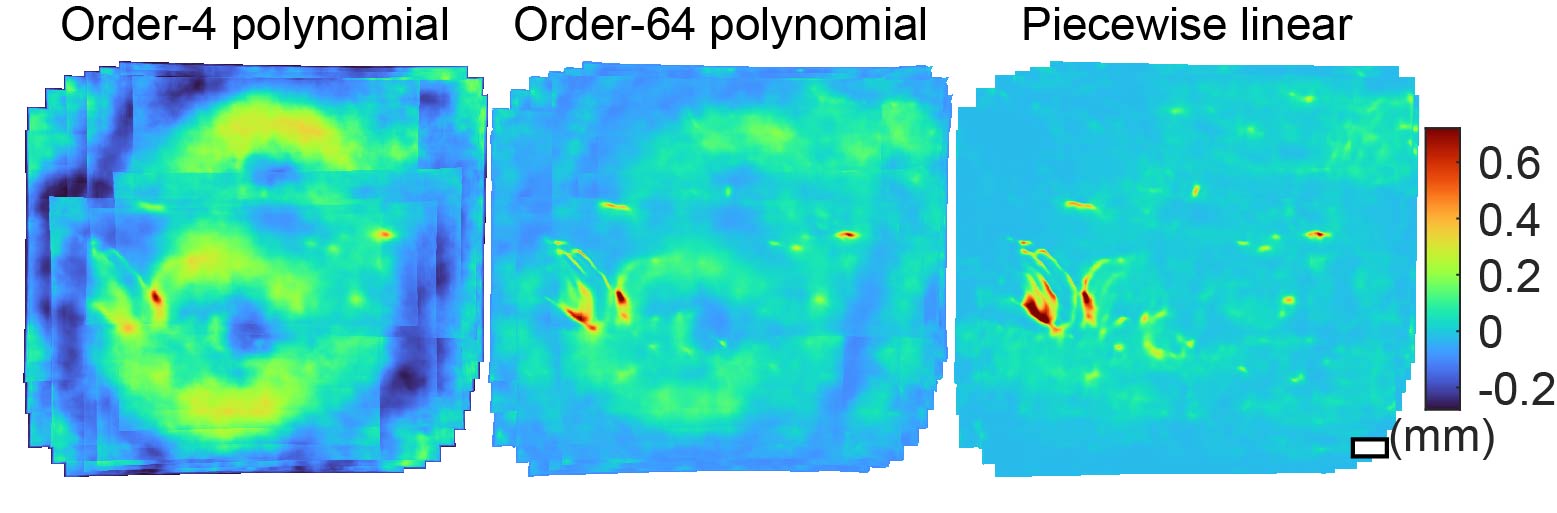}
    \caption{Strong ring artifacts in the painting sample due to uncompensated distortion in polynomial models. Scale bar, 1 cm.}
    \label{fig:distortion_painting}
\end{figure}
\begin{figure}
    \centering
    \includegraphics[width=\columnwidth]{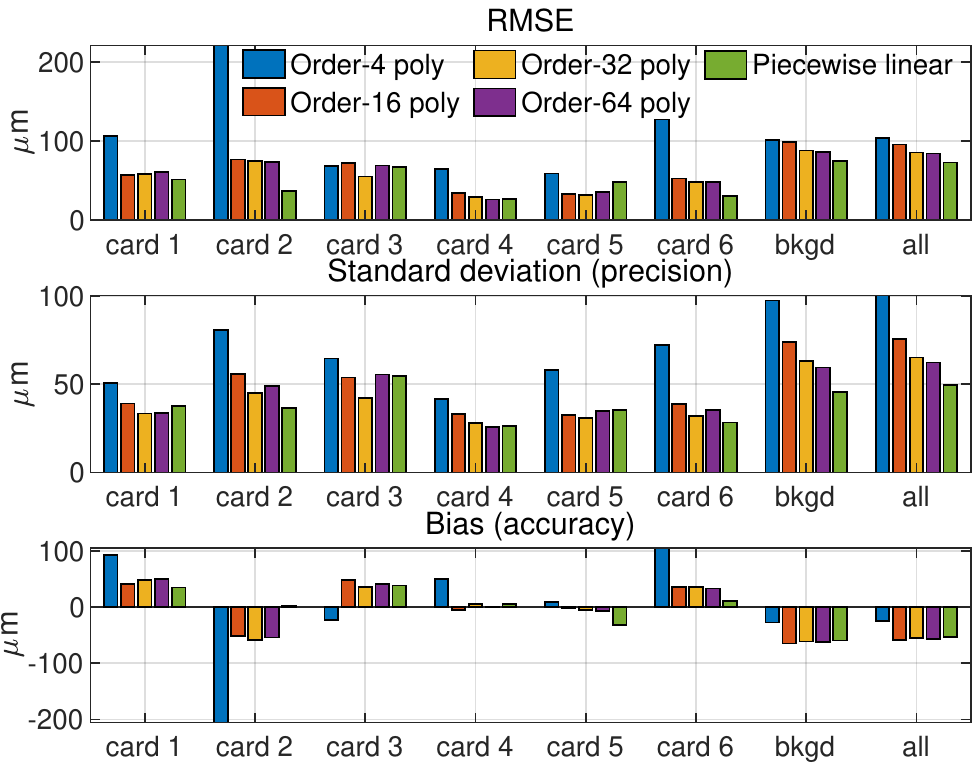}
    \caption{Height estimation performance on the 6 cut cards, background, and whole image for various undistortion models.}
    \label{fig:distortion_barplot}
\end{figure}
\begin{figure*}[ht]
    \centering
    \includegraphics[width=.9\textwidth]{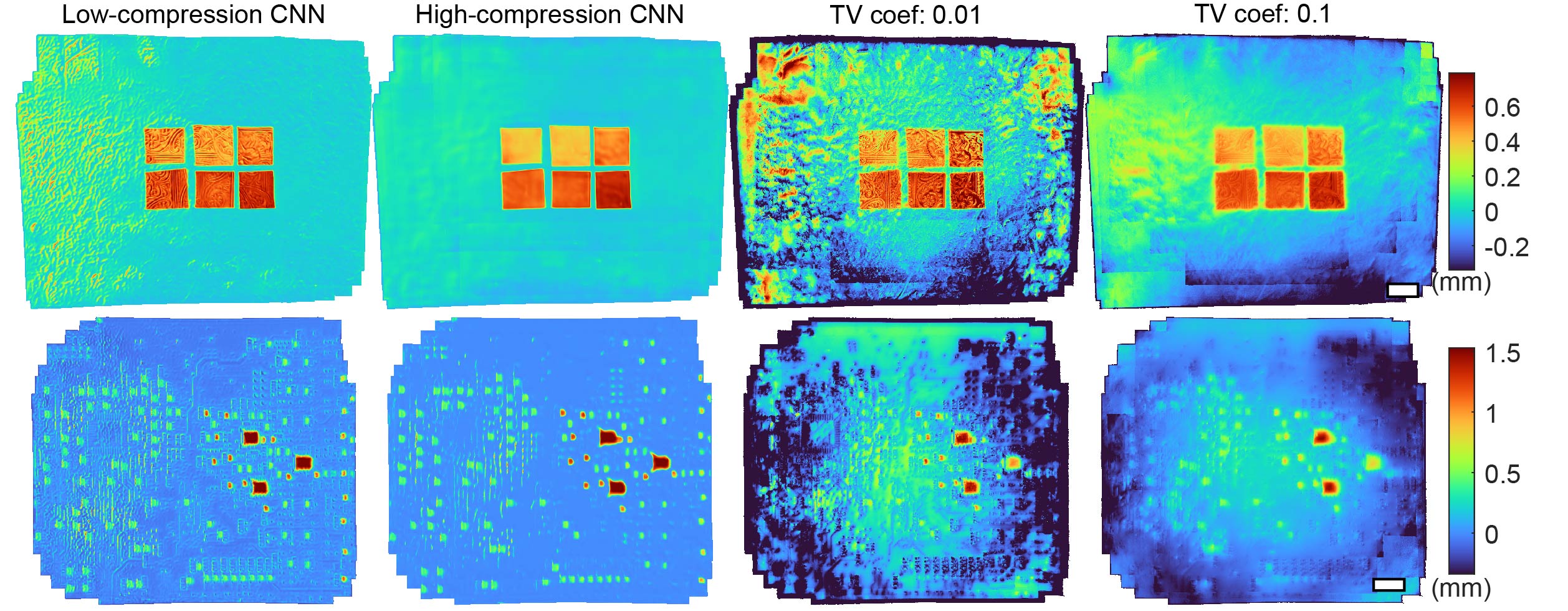}
    \caption{Reconstructions of the cut cards and PCB samples using high- and low-compression CNNs (architectures 1 and 4; see supplement for architectures and additional results) and high and low levels of TV regularization. Zoom in to see fine features. Scale bars, 1 cm.}
    \label{fig:reg_comp}
\end{figure*}

\noindent
\textbf{Accuracy and precision characterization.}
We first created a calibrated phantom sample consisting of standard playing cards ($\sim$0.3-mm thick), cut into six 1-2-cm\textsuperscript{2} squares and attached 0-5 layers of tape (50-70 \textmu m thick per layer) to alter their heights. We measured the thicknesses of the tape-backed cut cards using calipers with 20-\textmu m accuracy (Starrett EC799A-12/300), and arranged them on a flat, feature-rich surface \cite{bengio2017deep}. We regarded these measurements as the ground truths, $\mathbf{h_{gt}}$ (column 1 of Table \ref{card_analysis}).

The stitched orthomosaics and height map reconstructions by our method and COLMAP are shown in the first column of Fig. \ref{fig:results}, noting that COLMAP underestimates the card heights. To quantify accuracy and precision of both methods, we manually segmented the six cut cards and background based on the RGB orthomosaic, and computed the mean and standard deviation of the height values of these seven regions (Table \ref{card_analysis}). Because the height maps have an arbitrary global shift, we used the shift that minimizes the MSE between the mean height estimates and the ground truths: $\Delta h = mean(\mathbf{h_{gt}}-\mathbf{h_{est}})$. While COLMAP underestimates heights and therefore has low absolute accuracy, we hypothesized that the relative accuracy might be high. To test this, we additionally scaled COLMAP's height estimates by the factor that minimizes MSE between mean height estimates and ground truths, given by $cov(\mathbf{h_{gt}}, \mathbf{h_{est}})/var(\mathbf{h_{est}})\approx2.65$ (Table \ref{card_analysis}). Only our method has simultaneously high accuracy (26.3 \textmu m) and precision (37.8 \textmu m) and without the need to rescale.

\noindent
\textbf{Results on multiple mm-scale samples.}
Next, we compared our method and COLMAP on a printed circuit board (PCB), helicopter seeds, and the brush strokes on a painting (Fig. \ref{fig:results}). Unlike our method, COLMAP not only consistently underestimates heights, but also exhibits unpredictable surface curvatures. We quantified the height accuracy and precision on 16 of the PCB components using caliper estimates as the ground truth and manual segmentation based on the RGB orthomosaics (Table \ref{pcb_analysis}). We caution that this analysis is less rigorous than that of the cut cards sample, as some components are not completely flat (slight elevations along the shorter edges), and due to greater difficulty in estimating height with the calipers (we alleviated this problem statistically by using the mean of 10 measurements as the ground truth). These caveats may explain why the accuracy (62.5 \textmu m) and precision estimates (95.8 \textmu m) are inflated compared to those of the cut cards sample. However, it is clear that our method does not have COLMAP's underestimation issue. Further evidence of this is that the optimal rescale value for our method is $1.028\approx1$ (which we did not use). While we do not have ground truths for the helicopter or painting samples, the calipers estimated the helicopter seeds to be $\sim$2.3 mm at their thickest part, qualitatively consistent with our method; while this is not a rigorous measurement due to non-rigidity of the sample, it at least suggests that COLMAP is underestimating heights. 

\noindent
\textbf{Importance of undistortion.} Figs. \ref{fig:distortion_cards} and \ref{fig:distortion_painting} show spurious rings when camera distortions are not sufficiently modeled. In particular, although for conventional macro-scale applications it is often sufficient to use a low-order even polynomials (e.g., order-4 \cite{gordon2019depth}), for mesoscopic applications, even using an order-64 polynomial undistortion model leaves behind noticeable ring artifacts, while our piecewise linear model (30 segments) effectively eliminates them. Fig. \ref{fig:distortion_barplot} quantifies the performance of multiple undistortion models via root-MSE (RMSE), precision, and accuracy of heights of the cut cards sample. Not only does our piecewise linear model generally have better precision and overall RMSE than the polynomial models, but also it does not exhibit directional biases caused by the ring artifacts (e.g., card 2), which cannot be corrected by a global scale or shift. See the supplement for a full comparison of undistortion models on all four samples in Fig. \ref{fig:results}, as well as the estimated radial undistortion profiles and distortion centers.

\noindent
\textbf{Effectiveness of CNN regularizer.} CNN reparameterization is crucial to our method. As the CNN is an encoder-decoder network without skip connections, the degree of compression can be adjusted by the number of parameters and downsampling layers. Fig. \ref{fig:reg_comp} shows that varying the degree of compression controls the amount of fine detail transferred to the height map without affecting the flatness of the field of view or blurring edges. Thus, the degree of compression in the CNN is an interpretable means of tuning the regularization. However, if we optimize the camera-centric height maps directly with total variation (TV) regularization, we see many artifacts, even when the regularization is strong enough to blur sharp edges. See the supplement for results with more CNN architectures and TV regularization levels for all four samples in Fig. \ref{fig:results}, plus the hyperparameter-tuning sample.

\section{Conclusion}
We have presented a feature-free, end-to-end photogrammetric algorithm applied to mesoscopic samples with tens-of-\textmu m accuracy over cms fields of view. Our method features a novel use of CNNs/DIPs, which effectively removes many artifacts from the height maps. We also showed that careful modeling of distortion is important for obtaining accurate height values. Qualitative and quantitative comparisons show that our method outperforms COLMAP, a feature-based SfM tool. Although we applied our method to mesoscopic samples with the orthorectification model, it would be interesting to apply this to macro-scale scenes where photogrammetry is more typically applied, using the arbitrary reference rectification model (Fig. \ref{fig:camera_model}d). 

{\small
\bibliographystyle{ieee_fullname}
\bibliography{egbib}
}

\end{document}


\maketitle

\section{Homographic rectification}

Here, we describe the procedure for 2D backprojection of a point in the camera image plane in the camera's intrinsic coordinate system relative to the projection center, $\mathbf{r_{im}^{cam}}=(x_\mathit{im}^\mathit{cam}, y_\mathit{im}^\mathit{cam}, f_\mathit{ph})$, onto the object plane in the world reference frame. This procedure ignores 3D height variation, which are accounted for in a separate step (see main text). For clarity, $\mathit{cam}$ superscripts indicates that the variables are defined in the camera's reference frame, and $\mathit{obj}$ and $\mathit{im}$ subscripts reference the object and image planes, respectively.
\begin{enumerate}
    \itemsep0em
    \item Rotation in the image plane by $\theta$:
    \begin{equation} \label{step1}
        (x_\mathit{im}^\mathit{cam}, y_\mathit{im}^\mathit{cam})\leftarrow(x_\mathit{im}^\mathit{cam}, y_\mathit{im}^\mathit{cam})R(\theta),
    \end{equation}
    where $R(\theta)$ is a 2D rotation matrix. Let $\mathbf{r_{im}^{cam}}$ be accordingly updated.
    
    \item Backprojection to the object plane:
    \begin{equation}
        \mathbf{r_{obj}^{cam}}=\frac{Z}{\mathbf{\hat{n}_{obj}^{cam}} \cdot \mathbf{r_{im}^{cam}}}\mathbf{r_{im}^{cam}},
    \end{equation}
    where $\cdot$ denotes a dot product and $\mathbf{\hat{n}_{obj}^{cam}}=(n_x,n_y,n_z)$ is the unit normal vector of the object plane defined in the camera's reference frame. For example, $\mathbf{\hat{n}_{obj}^{cam}}=(0,0,-1)$ when the image and object planes are parallel.
    
    \item Coordinate change via 3D rotation from camera coordinates, $\mathbf{r_{obj}^{cam}}$, to world coordinates, $\mathbf{r_{obj}}=(x_\mathit{obj},y_\mathit{obj})$ (i.e., by angle, $\cos^{-1}(-n_z)$, about axis, $(-n_y, n_x, 0)$):
    \begin{equation}\label{backproject}
    \begin{gathered}
        x_\mathit{obj}=\frac{Z(1+n_z)\left((n_y^2+n_z-1)x_\mathit{im}-n_xn_yy_\mathit{im}\right)}{n_z(n_x^2+n_y^2)\mathbf{\hat{n}_{obj}^{cam}} \cdot \mathbf{r_{im}^{cam}}},\\
        y_\mathit{obj}=\frac{Z(1+n_z)\left((n_x^2+n_z-1)y_\mathit{im}-n_xn_yx_\mathit{im}\right)}{n_z(n_x^2+n_y^2)\mathbf{\hat{n}_{obj}^{cam}} \cdot \mathbf{r_{im}^{cam}}}.
    \end{gathered}
    \end{equation}
    In practice, this equation is numerically unstable, as it involves dividing $1+n_z$ by $n_x^2+n_y^2$, which are both 0 when the image and object planes are parallel. We instead use its second-order Taylor expansion at $n_x=0, n_y=0$, which is valid by our lateral-translation-dominant assumption ($|n_x|,|n_y| \ll |n_z|\approx1$).

\begin{align}
        x_\mathit{obj}\approx \frac{Z}{f_\mathit{ph}n_z}\biggl(&x_\mathit{im} + \frac{x_\mathit{im}(n_xx_\mathit{im}+n_yy_\mathit{im})}{f_\mathit{ph}} +\notag\\
        &\frac{f_\mathit{ph}^2n_x(n_xx_\mathit{im} + n_yy_\mathit{im}) + 2x_\mathit{im}(n_xx_\mathit{im} + n_yy_\mathit{im})^2}{2f_\mathit{ph}^2}
        \biggr),\\
        y_\mathit{obj}\approx \frac{Z}{f_\mathit{ph}n_z}\biggl(&y_\mathit{im} + \frac{y_\mathit{im}(n_xx_\mathit{im}+n_yy_\mathit{im})}{f_\mathit{ph}} + \notag\\
        &\frac{f_\mathit{ph}^2n_x(n_xx_\mathit{im} + n_yy_\mathit{im}) + 2y_\mathit{im}(n_xx_\mathit{im} + n_yy_\mathit{im})^2}{2f_\mathit{ph}^2}
        \biggr).
    \end{align}
Note that the zero-order terms correspond to the usual camera-centric perspective projection expressions.
    
    \item Addition of camera lateral position, $\mathbf{R}=(X,Y)$:
    \begin{equation}\label{step4}
        x_\mathit{obj}\leftarrow x_\mathit{obj}+X,\
        y_\mathit{obj}\leftarrow y_\mathit{obj}+Y.
    \end{equation}
\end{enumerate}
This backprojection procedure onto a common object plane is done for each camera image.

\section{COLMAP \cite{schonberger2016structure} hyperparameters}
We used the OpenCV fisheye camera model, shared among all images in the same sequences, with the focal length supplied via EXIF and the principal point (i.e., the projection center position) optimizable, as in our method. Exhaustive feature matching and guided matching were enabled, and all image-size-related settings were set so that COLMAP maintained the $1512\times2016$ input size. Finally, among Delaunay meshing hyperparameters, we set max\_proj\_dist=1 and quality\_regularization=2. The camera model and meshing hyperparameters were tuned using the same independent sample we used to select the CNN architecture. For all other hyperparameters, we used the defaults.

\section{Camera undistortion estimates}
\begin{figure}
    \centering
    \includegraphics[width=.6\textwidth]{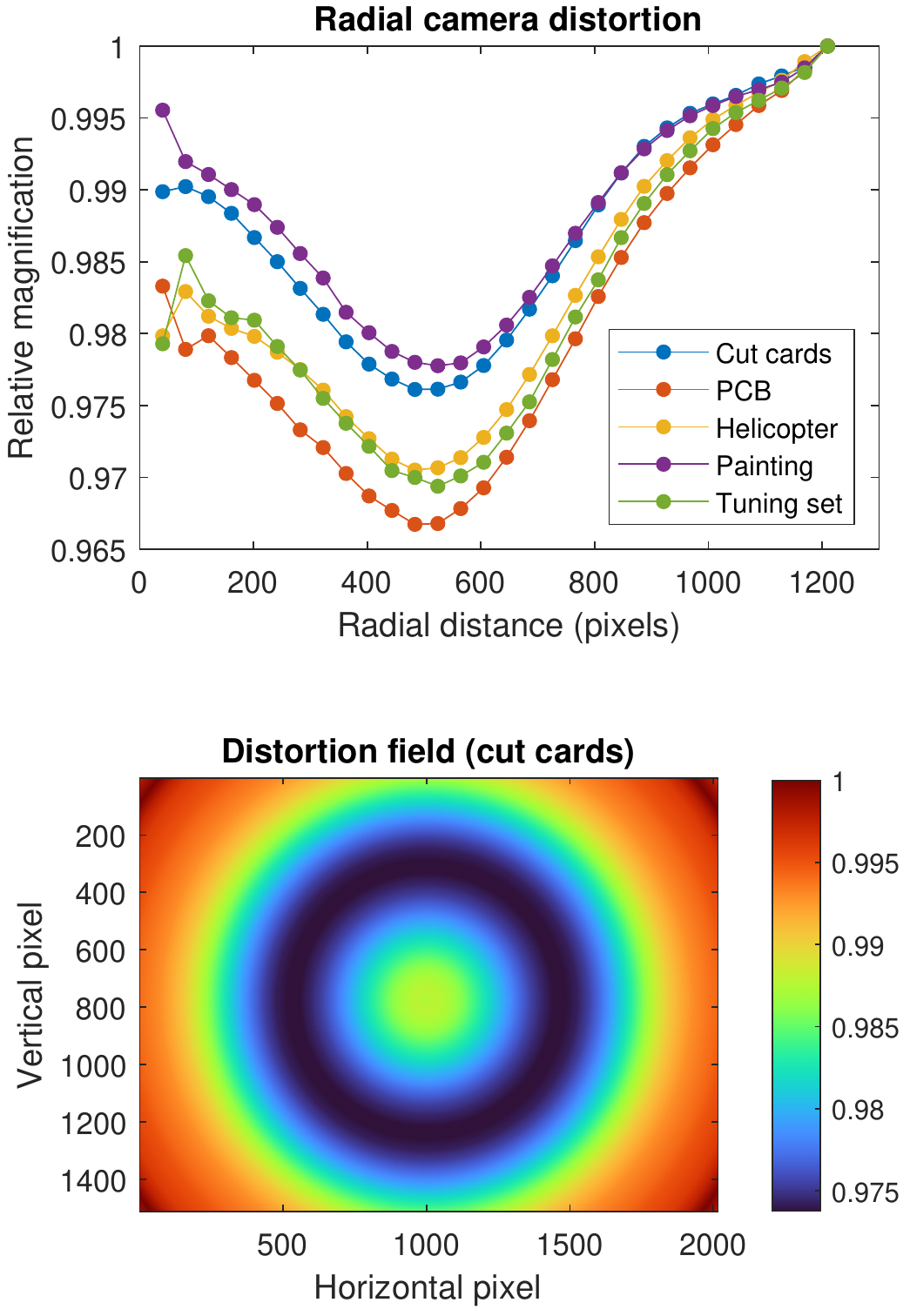}
    \caption{Top: radial undistortion for each sample. Bottom: sample 2D undistortion profile estimated using the cut cards sample. The 2D undistortion profiles estimated from the other samples are similar.}
    \label{fig:distortion}
\end{figure}

As mentioned in the main text, it is very important to correct for camera distortion, which are position-dependent relative magnifications. Fig. \ref{fig:distortion} shows the radial camera undistortion estimates for each sample under the piecewise linear model. Since there can be an arbitrary constant global magnification, in our implementation we arbitrarily normalized distortions to their maximum values. This procedure does not constrain our results, as an error in global magnification can be compensated by a global height shift, and vice versa. We also allowed the camera principal point (i.e., the projection center position) to vary; the results are shown in Table \ref{tab:distortion_center} for all four samples in the main text plus the hyperparameter tuning sample. Note that the units are in pixels \textit{after} we downsampled the original camera images by $2\times$. 

Overall, the results are qualitatively consistent across different samples, with small deviations due to reconstruction error and/or changing distortion properties as a function of sensor or lens repositioning for autofocus.

\begin{table}[]
    \centering
    \begin{tabular}{c|c|c}
         Sample & Vert. & Horiz. \\
         \hline
         Cut cards & 17.5 & -9.6\\
         PCB & 20.2 & -20.0\\
         Helicopter & 19.8 & -14.6\\
         Painting & 17.0 & -15.5\\
         Tuning sample & 22.3 & -17.9
    \end{tabular}
    \caption{Estimates of the camera principal point position relative to the center (pixels).}
    \label{tab:distortion_center}
\end{table}

\section{Full comparison of undistortion models}
\begin{figure}
    \centering
    \centerline{\includegraphics[width=1.6\textwidth]{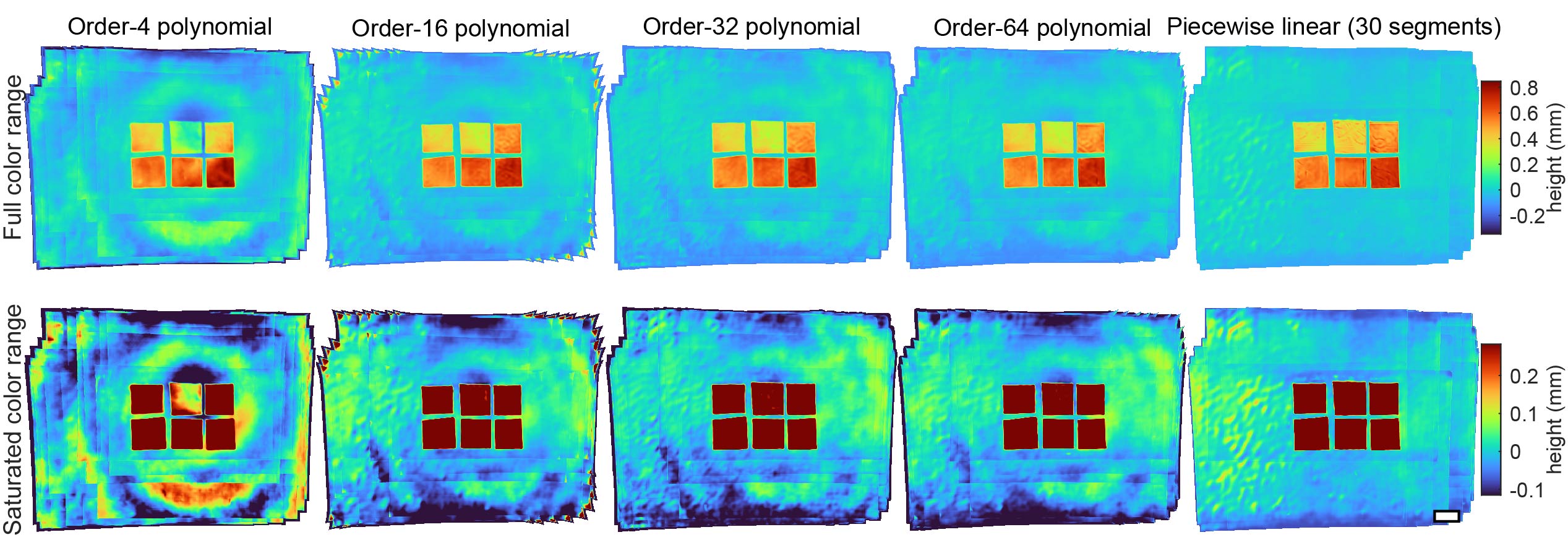}}
    \caption{Comparison of height map reconstructions of the cut cards sample under various undistortion models. The second row uses a saturated color range to emphasize artifacts. Scale bar, 1 cm. }
    \label{fig:cut_cards_distort}
\end{figure}

\begin{figure}
    \centering
    \centerline{\includegraphics[width=1.6\textwidth]{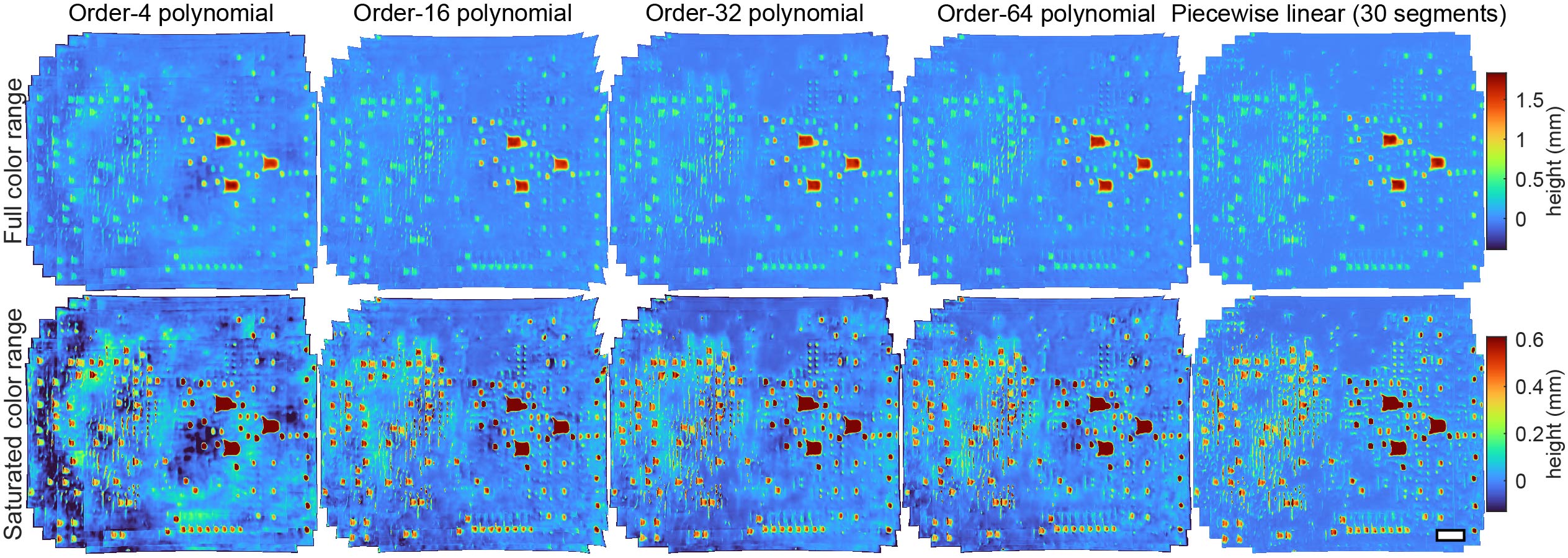}}
    \caption{Comparison of height map reconstructions of the PCB sample under various undistortion models. The second row uses a saturated color range to emphasize artifacts. Scale bar, 1 cm. }
    \label{fig:PCB_distort}
\end{figure}

\begin{figure}
    \centering
    \centerline{\includegraphics[width=1.6\textwidth]{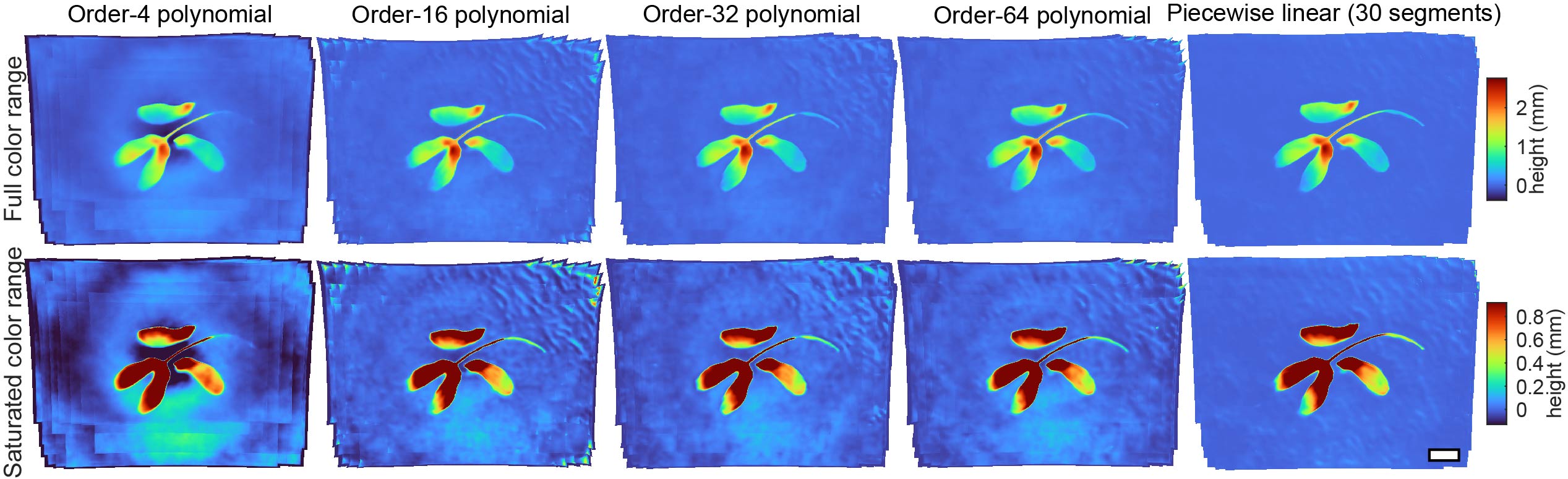}}
    \caption{Comparison of height map reconstructions of the helicopter seeds sample under various undistortion models. The second row uses a saturated color range to emphasize artifacts. Scale bar, 1 cm. }
    \label{fig:helicopter_distort}
\end{figure}

\begin{figure}
    \centering
    \centerline{\includegraphics[width=1.6\textwidth]{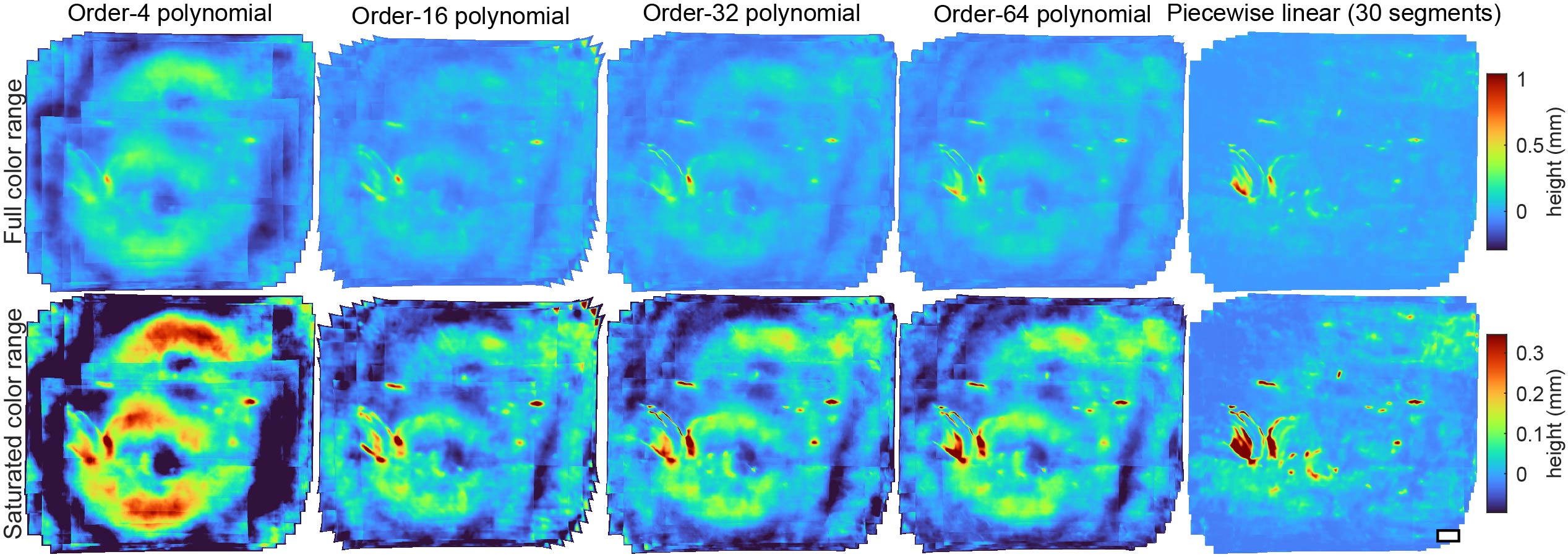}}
    \caption{Comparison of height map reconstructions of the painting brush strokes sample under various undistortion models. The second row uses a saturated color range to emphasize artifacts. Scale bar, 1 cm. }
    \label{fig:painting_distort}
\end{figure}

Here, we show the full comparison of height map reconstructions on all four samples using a 4th, 16th, 32nd and 64th order even polynomials (2, 8, 16, and 32 coefficients, respectively) as well as our piecewise linear radial undistortion model (with 30 line segments). These results are shown in Figs. \ref{fig:cut_cards_distort}-\ref{fig:painting_distort}. The second rows of these figures have saturated color ranges to better appreciate the distortion-induced artifacts, which typically manifest as rings and affect the polynomial models, but not the piecewise linear model. 

\section{CNN architectures}
We use slightly modified versions of the encoder-decoder CNNs (no skip connections) from the original DIP paper \cite{ulyanov2018deep}. Specifically, we composed CNNs using the downsample and upsample blocks summarized in Table \ref{tab:blocks}. We designed different symmetric architectures by varying the number of upsample/downsample pairs and number of filters, $k$, per block. For example, the architecture specified by the list, $[16, 32, 64]$, consists of three sequential downsample blocks with $k=16, 32, \text{and}\ 64$ filters, respectively, followed by three sequential upsample blocks with $k=64, 32, \text{and}\ 16$ filters. For all reconstructions in the main text, we used $[16, 16, 16, 32, 32]$, which we arrived at by comparing reconstruction quality on an independent sample (see next section).

\begin{table}[]
    \centering
    \begin{tabular}{c|c}
         Downsample block & Upsample block\\
         \hline
         & $2\times$ bilinear upsample\\
         $3\times3$ conv, $k$ filters, stride=2& $3\times3$ conv, $k$ filters, stride=1\\
         Batch normalization & Batch normalization\\
         Leaky ReLU & Leaky ReLU\\
         $3\times3$ conv, $k$ filters, stride=1&$1\times1$ conv, $k$ filters, stride=1 \\
         Batch normalization & Batch normalization\\
         Leaky ReLU & Leaky ReLU\\
         
    \end{tabular}
    \caption{Upsample and downsample blocks used in our encoder-decoder architectures. Convolutions use `same' padding mode.}
    \label{tab:blocks}
\end{table}

\section{Full comparison of CNN-based and TV regularization}
Here, we show height map reconstructions for all four samples in the main text, as well as the independent hyperparameter-tuning sample, using four different CNN architectures:
\begin{itemize}
\item Architecture 1: $[16, 16, 32, 32]$, 69,424 parameters
\item Architecture 2: $[16, 16, 16, 16]$, 28,080 parameters
\item Architecture 3: $[16, 16, 16, 32, 32]$, 76,912 parameters (used in main text)
    \item Architecture 4: $[16, 16, 16, 16, 16]$, 35,568 parameters
\end{itemize}
The motivation for these choices was to test two different methods of compression in the CNN -- using fewer parameters and using more downsampling blocks. Note that in all cases, the number of parameters is significantly fewer than the number of camera-centric height map pixels (60-70 million, depending on number of images in the sequence). We chose architecture 3 for all the reconstruction figures in the main text (unless otherwise specified), because it balances loss of resolution and reduction of artifacts in the background of the tuning sample.

We also show four different levels of isotropic total variation (TV) regularization,
\begin{equation}
    TV(h(x,y))=\sum_{x,y}{\sqrt{|\nabla_xh(x,y)|^2+|\nabla_yh(x,y)|^2}},
\end{equation}
where the directional gradients are approximated by finite differences. The total loss is therefore MSE + $\lambda$TV, where we used $\lambda=0.003, 0.01, 0.03, \text{and}\ 0.1$.

\begin{figure}[!hbp]
    \centering
    \centerline{\includegraphics[width=1.6\textwidth]{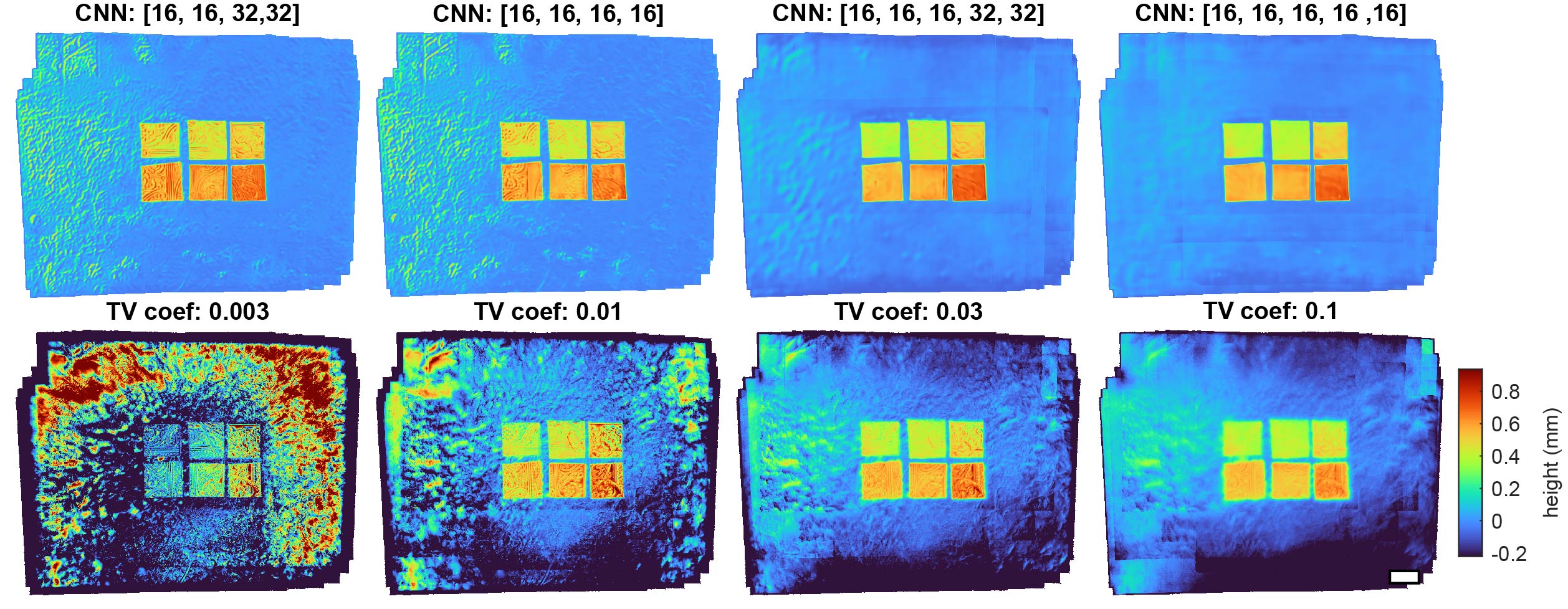}}
    \caption{Regularization comparison for the cut cards sample. Scale bar, 1 cm.}
    \label{fig:cards}
\end{figure}

\begin{figure}
    \centering
    \centerline{\includegraphics[width=1.6\textwidth]{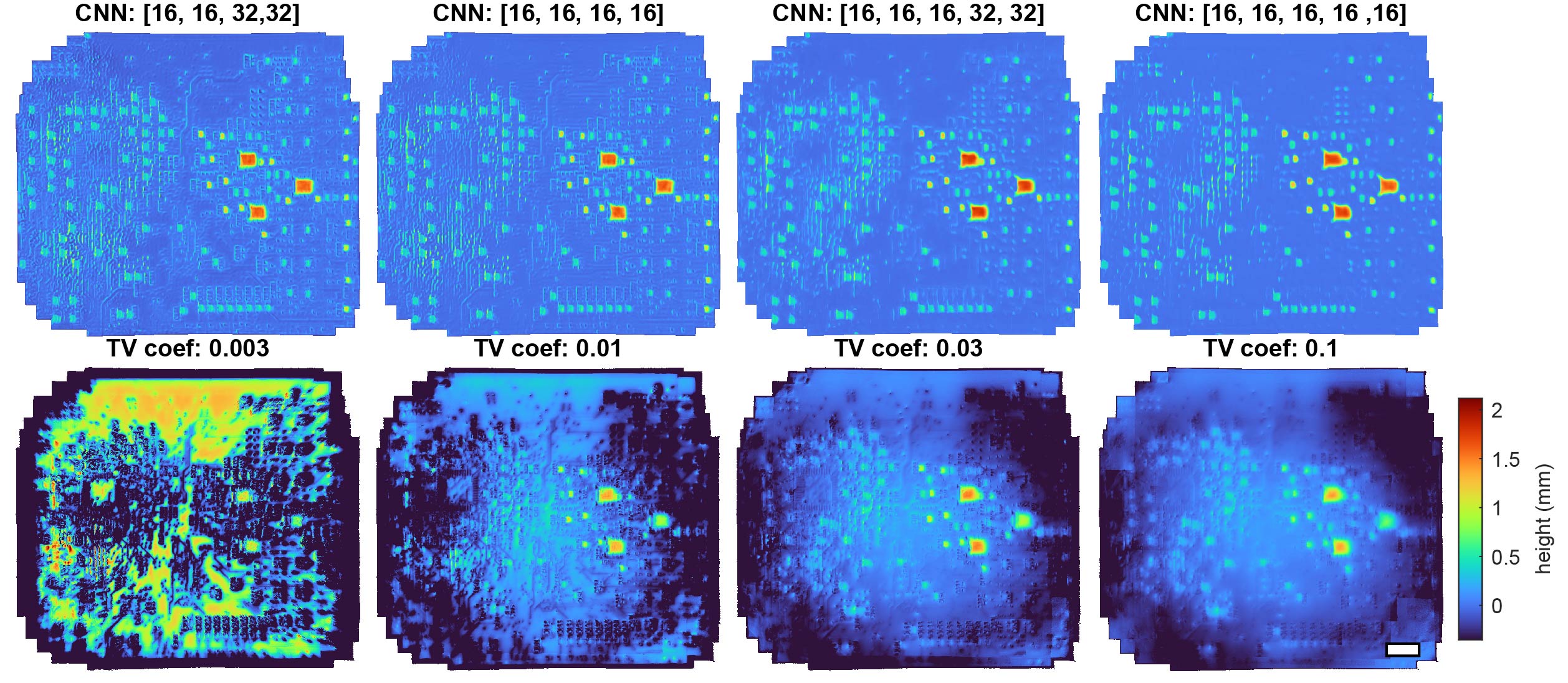}}
    \caption{Regularization comparison for the PCB sample. Scale bar, 1 cm.}
    \label{fig:pcb}
\end{figure}

\begin{figure}
    \centering
    \centerline{\includegraphics[width=1.6\textwidth]{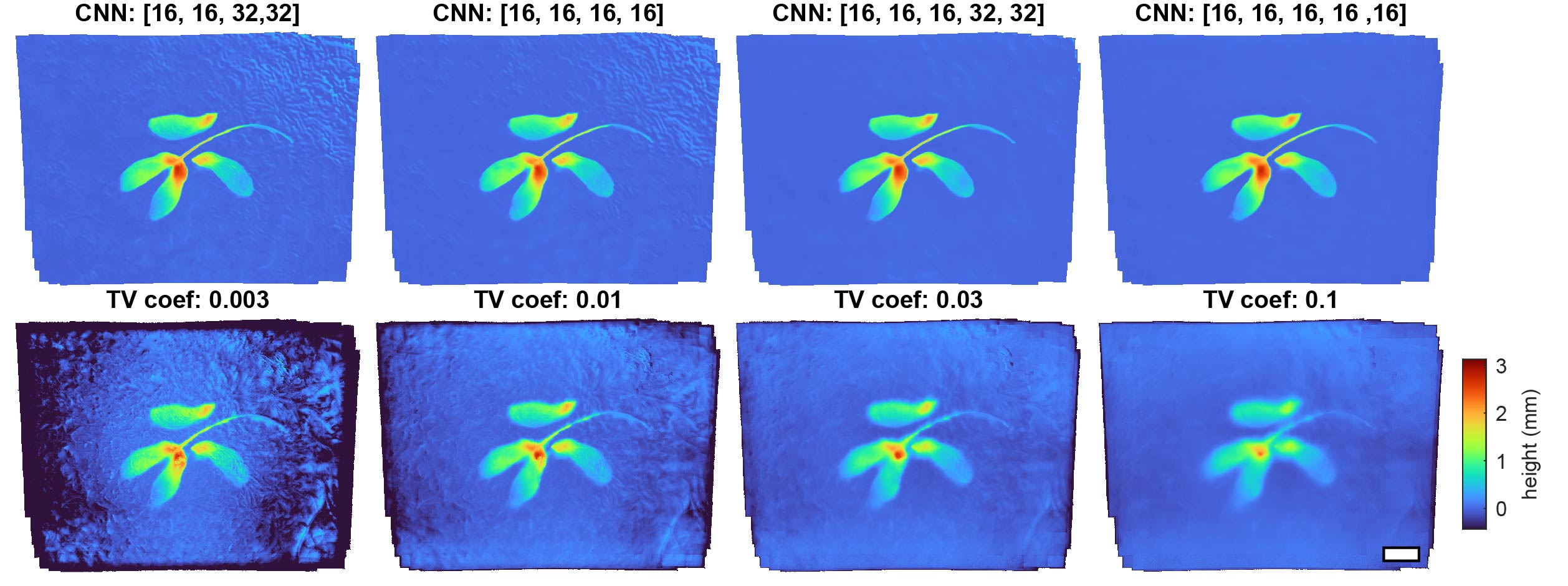}}
    \caption{Regularization comparison for the helicopter sample. Scale bar, 1 cm.}
    \label{fig:helicopter}
\end{figure}

\begin{figure}
    \centering
    \centerline{\includegraphics[width=1.6\textwidth]{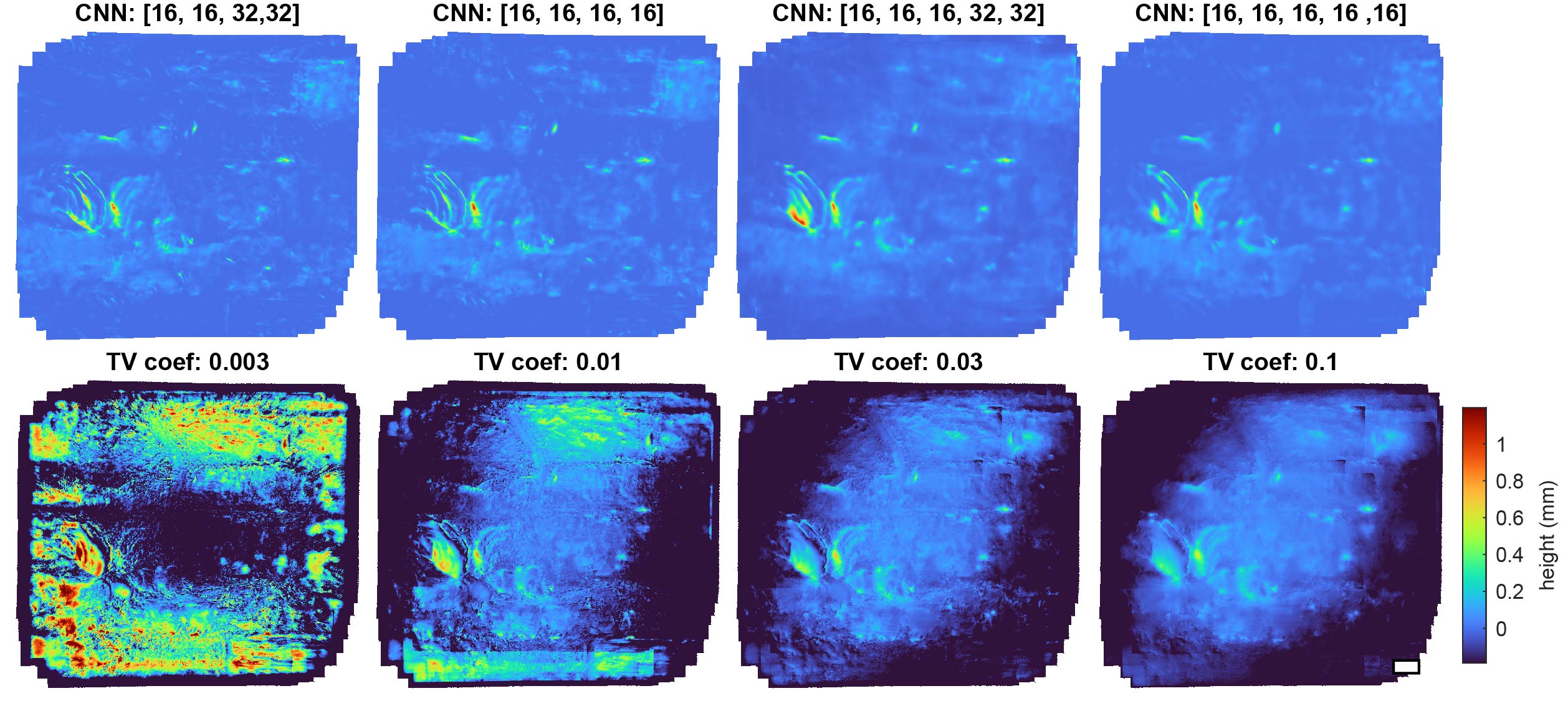}}
    \caption{Regularization comparison for the painting brush strokes sample. Scale bar, 1 cm.}
    \label{fig:painting}
\end{figure}

\begin{figure}
    \centering
    \centerline{\includegraphics[width=1.6\textwidth]{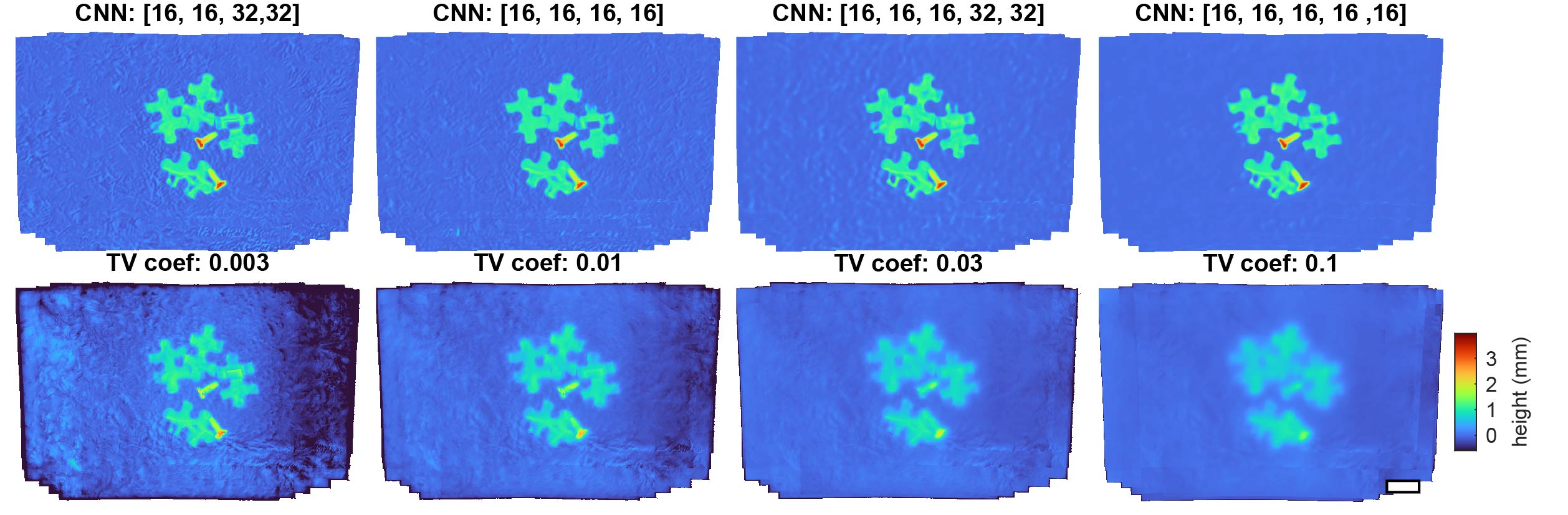}}
    \caption{Regularization comparison for the hyperparameter tuning sample (puzzle pieces and screws). The third CNN architecture was chosen to balance resolution and reduction of artifacts in the background. Scale bar, 1 cm.}
    \label{fig:hyper}
\end{figure}

The comparisons are show in Figs. \ref{fig:cards}-\ref{fig:hyper}. All height reconstructions within the same figure share the same color range. In general, the compression-based CNN regularization has more desirable properties as the regularization strength is tuned, such as flat backgrounds regardless of regularization strength.

{\small
\bibliographystyle{ieee_fullname}
\bibliography{egbib}
}


